\documentclass{article}

\usepackage[final]{neurips_2021}
\usepackage{environ}

\usepackage[utf8]{inputenc} 
\usepackage[T1]{fontenc}    
\usepackage{hyperref}       
\usepackage{url}            
\usepackage{booktabs}       
\usepackage{amsfonts}       
\usepackage{nicefrac}       
\usepackage{microtype}      
\usepackage{xcolor}         

\usepackage{graphicx}
\usepackage{subfigure}

\usepackage{tabularx}
\usepackage{multicol}
\usepackage{multirow}
\usepackage{amsmath}
\usepackage{cleveref} 
\usepackage{mathrsfs}
\usepackage{amsthm}
\usepackage{stackengine}
\usepackage{algorithm}
\usepackage{algorithmic}

\title{A Binded VAE for Inorganic Material Generation}

\author{
      Fouad Oubari\\
      Centre Borelli, UMR 9010, ENS Paris Saclay, Michelin\\
      \texttt{fouad.oubari@ens-paris-saclay.fr} \\
   \And
       Antoine De Mathelin \\
       Centre Borelli, UMR 9010, ENS Paris Saclay, Michelin\\
       \texttt{antoine.de-mathelin-de-papigny@michelin.com} \\
   \AND
      Rodrigue Décatoire \\
      Michelin \\
      \texttt{rodrigue.decatoire@michelin.com} \\
   \And
      Mathilde Mougeot \\
      Centre Borelli, UMR 9010, ENS Paris Saclay, ENSIIE \\
      \texttt{mmougeot@ens-paris-saclay.fr} \\
}
\begin{document}
	
\maketitle

\begin{abstract}
Designing new industrial materials with desired properties can be very expensive and time consuming. The main difficulty is to generate compounds that correspond to realistic materials. Indeed, the description of compounds as vectors of components' proportions is characterized by discrete features and a severe sparsity. Furthermore, traditional generative model validation processes as visual verification, FID and Inception scores are tailored for images and cannot then be used as such in this context. To tackle these issues, we develop an original Binded-VAE model dedicated to the generation of discrete datasets with high sparsity. We validate the model with novel metrics adapted to the problem of compounds generation. We show on a real issue of rubber compound design that the proposed approach outperforms the standard generative models which opens new perspectives for material design optimization.

\end{abstract}

\section{Introduction}
\label{intro}

The design of new drugs and materials with desired properties is an expensive and time consuming process. It traditionally involved laborious trial-and-error procedures. Evolutionary methods such as genetic algorithms \cite{lameijer2006molecule,chakraborti2004genetic} accelerated the process, for instance by modifying molecules structures to obtain molecules with desired properties. However, these methods need hand-crafted rules and constraints to avoid unrealistic generations, which might be hard or even impossible to express mathematically. 
To bypass this issue, a good alternative are deep generative methods that are able to learn the data space distribution without the need of constraint definition.

In the past few years, deep generative models such as GANs \cite{goodfellow2014generative} and VAEs \cite{Kingma2014AutoEncodingVB} have been successfully applied to different tasks ranging from classification to generation \cite{karras2019style, vahdat2020nvae} or even image translation \cite{zhu2017unpaired}.
However, as observed in our experiments (cf Section \ref{expes}), these two models do not provide satisfactory solutions for the generation of new materials. The difficulties encountered are the 
severe sparsity of the data combined with the diversity of components' ratios from one compound to another. Moreover, GANs and VAEs require their generations to be differentiable and thus fail to generate discrete compounds \cite{gui2020review}. Their inability to directly generate discrete data appears to be an open problem.


Furthermore, unlike when generating images, visual inspection of each generated compound is more complex and requires an expert's validation. Finally, the traditional metrics used to evaluate a generative model are based on the internal features of a deep network trained on ImageNet \cite{krizhevsky2012imagenet} as the Inception network \cite{Szegedy2016InceptionNetwork}; therefore, they cannot be used as such for material design.

To overcome all the above-mentioned issues, several works are carried out in the field of new compound generation \cite{ maziarka2020mol, guimaraes2017objective, kusner2017grammar,gomez2018automatic}. For instance when working with GANs for inorganic material generation, the Negative Dice Loss \cite{shamir2019continuous} is adopted to enforce the discontinuity of the generations by enforcing the overlap between the real and the generated data \cite{dan2020generative}. Our work is in line with this research field and proposes the two following main contributions:
\begin{itemize}
    \item First, to deal with the sparsity and the diversity of components' ratios, we introduce a Binded-VAE model (BVAE) which learns to jointly generate the raw data mask (a binary vector encoding whether or not a component is present in the compound) and the ratio of the components in the compound description (cf Section \ref{bvae}).
    \item 
    Second, we propose three metrics tailored to the problem of compound generation. The first one is an adaptation of the FID. The second one is the KL divergence computed between the distributions of real compound properties and predictions of these properties for generated compounds. The last one is a metric based on the agreement of "expert rules" (cf Section \ref{metrics}).
    \end{itemize}
    

We show, on a real dataset of rubber compound, that the proposed BVAE provides outperforming results compared to the ones obtained with standard VAEs and GANs (cf Section \ref{expes}). Thus, the combination of our model with an optimization algorithm can be considered as a promising tool for industrials wishing to generate new compounds with desired properties. Furthermore, the developed evaluation metric based on "expert rules" agreement appears to be a suitable validation process for specific issues as industrial ones. We are convinced that this metric can easily be extended to other industrial problems as long as expert rules can be defined on the dataset.







\section{Methodology}

We consider the problem of compound generation, where the learner has access to a sample of $n$ real compounds with $p$ components, and denote $X \in \mathbb{R}_+^{n \times p}$ the corresponding design matrix. Each compound $x \in \mathbb{R}_+^{p}$ is represented by a $p$ multidimensional vector denoted by $x = (x_1, ..., x_p)$ where $x_i \in \mathbb{R}_+$ is the ratio of the $\text{i}^\text{th}$ component in the compound such that $\sum_{i=1}^n{x_i}=1$. In practice, each vector is very sparse and only a few $x_i$ are not null. The goal of this work is to build a model able to simulate a dataset of new compounds $\widetilde{X}$ distributed in the same way than $X$. We denote $\widetilde{x} = (\widetilde{x}_1, ..., \widetilde{x}_p)$ the multidimensional vector corresponding to $\widetilde{X}$. 

Because of the sparsity in the raw data $X$, we observe that a standard VAE fails to generate a compound with a few non-zero components and adequate ratios (cf section \ref{results}). In order to overcome this problem, we decide to split the learning of the binary mask and the ratios into a Binded-VAE model (BVAE) as described in the next section.
\subsection{BVAE model}
\label{bvae}
\begin{figure}[ht]
	\begin{center}
		\includegraphics[width=\textwidth]{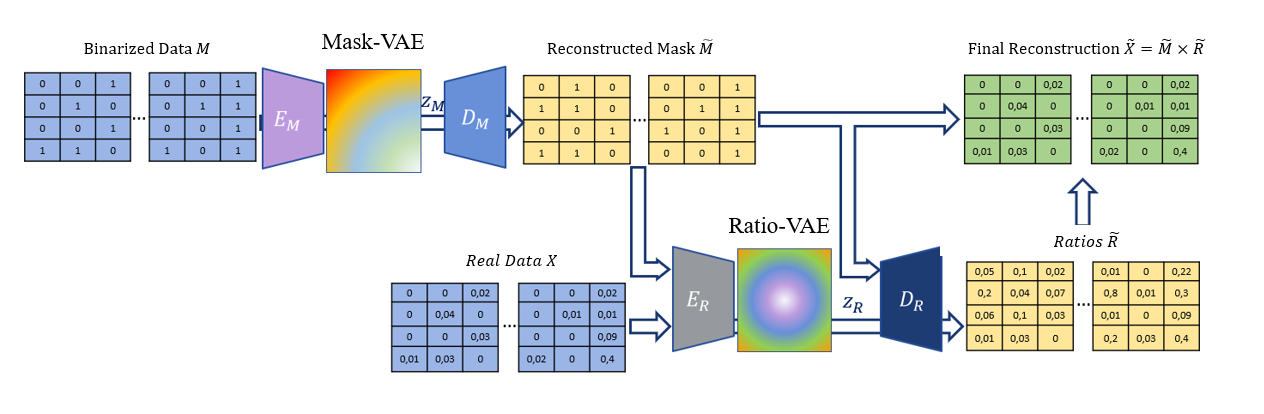}
		\caption{BVAE architecture. The mask $\widetilde{M}$ generated by the Mask-VAE contains the positions of the components in the generated compound. This mask is given as a condition to the Ratio-VAE, which then generates a vector $\widetilde{R}$ containing the ratio of each component.}
		\label{CVAE}
	\end{center}
\end{figure}

The BVAE is composed of two chained Variational AutoEncoders (cf Figure \ref{CVAE}). The first one, denoted "Mask-VAE" ($\text{VAE}_M$) receives as input a binary matrix $M \in \{0, 1\}^{n \times p}$ obtained by binarizing the real data $X$ such that $M_{ij} = 1$ if $X_{ij} > 0$ and $M_{ij} = 0$ elsewhere. It generates a reconstructed binary mask $\widetilde{M}$. The second VAE referred as  "Ratio-VAE" ($\text{VAE}_R$) is a Conditional VAE \cite{Sohn2015CVAE} that receives the concatenation of the real training data $X$ and its corresponding reconstructed mask $\widetilde{M}$ generated by the Mask-VAE. It generates a reconstructed set of ratios $\widetilde{R}$. The final reconstruction of $X$ is then obtained by multiplying the mask $\widetilde{M}$ with the output of the Ratio-VAE such that $\widetilde{X} = \widetilde{M} \times \widetilde{R}$. The final reconstruction $\widetilde{X}$ is differentiable so that the involved gradient can be calculated.  For the experiments in section \ref{expes} the generated data are truncated to the $4^{th}$ decimal place.

The loss of the Ratio-VAE is a classical VAE loss:
\begin{equation}
	\label{vae2_loss}
	\begin{split}
		\mathcal{L}_{\text{VAE}_R }= \lambda_R \, \text{BCE}(X, \widetilde{X}) + \text{KL}(\mathcal{N}(\mu_x,\sigma_x),\mathcal{N}(0,\,I))
	\end{split}
\end{equation}

Where $\text{BCE}$ is the Binary Cross-entropy defined as:
\begin{equation}
	\label{bce}
	\begin{split}
		\text{BCE}(X, \widetilde{X})= \frac{1}{n}\sum_{i=1}^n{\widetilde{x}_i\times\log(x_i)+(1-\widetilde{x}_i)\times\log(1-x_i)}
	\end{split}
\end{equation}
and $\text{KL}$ the Kullback-Leibler divergence. $\mu_x,\sigma_x$ are the outputs of the $\text{VAE}_R$ encoder.





For the Mask-VAE, the overlap between the generated mask $\widetilde{M}$ and the binarized input $M$ must be maximized. To this purpose, inspired by measures of regions overlap in image segmentation \cite{jadon2020survey}, we add to the classical VAE loss, the Negative Dice loss (NegDICE) \cite{shamir2019continuous} which has been recently adopted for inverse design of inorganic materials \cite{dan2020generative}. The NegDICE loss between two vectors $a, b \in [0,1]^p$ with $a = (a_1, ..., a_p)$ and $b = (b_1, ..., b_p)$ is defined as follows:
\begin{equation}
	\label{Shamir2017Dice}
		\text{NegDICE}(a, b) =  -2 \, \frac{a \cdot b}{||a||_1+||b||_1}=-2 \, \frac{\sum_{i=1}^p{a_ib_i}}{\sum_{i=1}^p{a_i} + \sum_{i=1}^p{b_i}}
\end{equation}





Therefore, the loss of the Mask-VAE is written as follows:
\begin{equation}
	 \label{vae1_loss}
		\mathcal{L}_{\text{VAE}_M} =\lambda_{M} \, \text{BCE}(M, \widetilde{M})) + \text{KL}(\mathcal{N}(\mu_x,\sigma_x), \mathcal{N}(0,I)) + \gamma \, \text{NegDICE}(M, \widetilde{M})
\end{equation}
Finally, the total loss of the BVAE is:
\begin{equation}
	\label{cvae_loss}
		\mathcal{L}_{\text{BVAE}} = \beta \, \mathcal{L}_{\text{VAE}_M} + \mathcal{L}_{\text{VAE}_R}
\end{equation}
 With $\lambda_R$,$\lambda_M$,$\gamma$ and $\beta$ positive parameters selected with gridsearch.

\subsection{Evaluation Metrics}

As mentioned in Section \ref{intro}, standard metrics used to evaluate generative models are tailored for images and thus are not suited for compound generation. We then propose the followings metrics:

\label{metrics}
\begin{itemize}
    \item \textbf{Custom FID (FID*).} The FID is the most widespread score used to evaluate the quality of the images of a generative model \cite{heusel2017gans} (cf equation \ref{fid}). The FID compares the first two moments of the real and generated distributions in an embedding given by the penultimate layer of a pretrained deep network. The FID score is then computed as follows: 
    \begin{equation}
	\label{fid}
		FID(R,G) = \left\|\mu_{R}-\mu_{G}\right\|+Tr(\Sigma_{R}+\Sigma_{G}-2(\Sigma_{R} \cdot \Sigma_{G})^\frac{1}{2})
    \end{equation}
    With $\mu_{R}$,$\mu_{G}$ and $\Sigma_{R}$,$\Sigma_{G}$ the first and the second moments of two distributions $R, G$. The deep network usually used to compute the FID is the Inception network \cite{Szegedy2016InceptionNetwork} trained on ImageNet \cite{krizhevsky2012imagenet}. This network is trained on images and is then not suited for our problem. Consequently, we adapt the computation of the FID to our problem by using a fully-connected neural network trained on the real compound dataset to predict properties of interest associated to each compound. We denote this custom metric FID*.
    
    \item \textbf{KL divergence.} Following the same idea, we use the aforementioned predictor to predict the properties for the generated data. We then compute the KL divergence between the predictor outputs of the generated data and the real distribution of the material properties.

    \item \textbf{"Expert Rules Agreement" metric (ERA).} 
    As our data represents compounds of real materials, they must verify specific rules given by experts as for instance, bounds on the total proportion of several components in the compound. We then define the ERA metric as the percentage of generated data that match the expert rules.
\end{itemize}

\section{Experiments}
\label{expes}
In our experiments, we compare different generative models trained on a real rubber compound dataset. In this chapter, we start by presenting the experimental setup, then the dataset used for training the models, and lastly we present and discuss the results obtained. 
\subsection{Setup}
\textbf{Baselines.} To perform our experiments we use different generative models: a GAN, a Vanilla VAE, our BVAE model, and an "Unbinded-VAE" model built with a "Mask-VAE" and a "Ratios-VAE" trained successively and separately. 

The models used in our experiments are composed of fully connected layers with the architectures given in Appendix \ref{architectures}. We use the Adam optimizer \cite{kingma2014adam} for all trainings.

Hyper-parameters selection is performed through gridsearch. The range for each hyper-parameter gridsearch are: [$10^{-1}, 1, 10]$ for $\lambda_R$,$\lambda_M$,$\gamma$ and $\beta$ (cf section \ref{bvae}), $[32, 64, 128, 256]$ for the latent dimensions and $[10^{-3}, 10^{-2}, 10^{-1}]$ for the learning rates.
The selected hyper-parameters are provided in Appendix \ref{architectures} Table \ref{tab: hyperparameters}.

\textbf{Dataset.} To train the models, we use a dataset of \textbf{rubber} compound of size $ \sim 12k$. The dataset is made of  1D vectors of components ratios, with a size of 99 and with an average sparsity of $ \sim 80\%$.
We train all the models for 200 epochs, with a batch size of 128. The evaluation metrics presented in Section \ref{metrics} are computed at each epoch.
The predictor used to compute the FID* and KL scores is trained using a dataset containing the values of three properties of interest for each compound.


\subsection{Results and Discussion}
\label{results}
In this section, we discuss the experiments results. We start by comparing the heatmaps of the generations and the real data; then, we compare the models with respect to the custom metrics introduced in \cref{metrics}. As mentioned in section \ref{bvae}, generations are truncated to the $4^{th}$ decimal place.

\begin{figure}[t]
	\begin{center}
	\includegraphics[width=\textwidth]{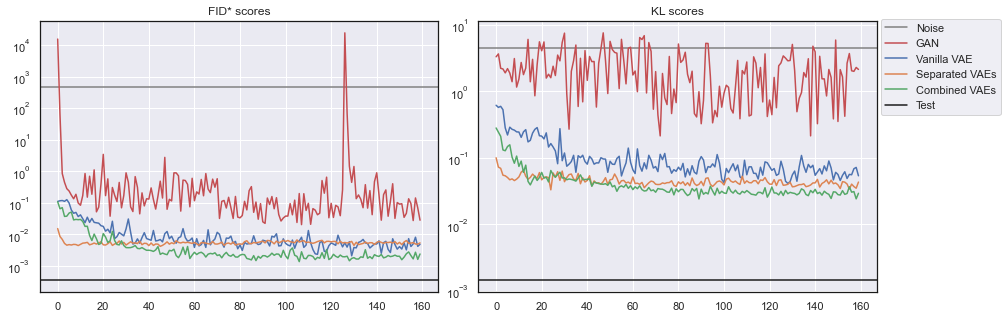}
	\caption{Evolution of FID* and KL scores by epoch. The two metrics are in coherence in terms of models scores. The test score is computed between two subsets of the real data, and the noise score is computed between the real data and a random Gaussian noise. }
	\label{fid_inception}
	\end{center}
\end{figure}

\textbf{Data visualization.}
\begin{figure}[t]
    	\begin{center}
		\includegraphics[width=\textwidth]{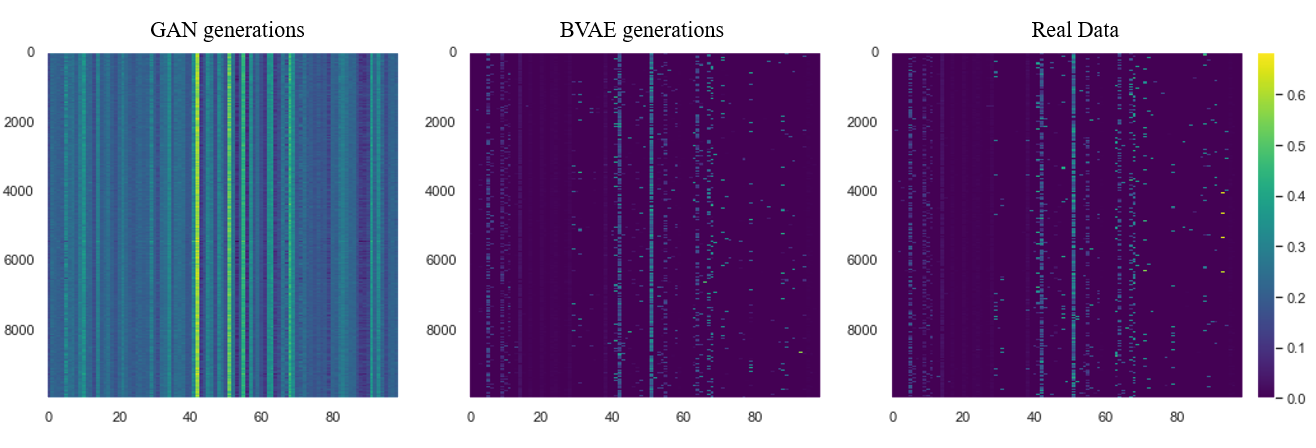}
		\caption{Generations and real data heatmaps. The 10 000 generations and real data vectors displayed are of size 99.}
		\label{generations}
	\end{center}
\end{figure}
Contrarily to image generation, inspecting each individual generated rubber compound would require an expert. However, the presence of patterns in the data provides a visual means to verify the generations. We propose to visualize the whole generated datasets as a 2D heatmap where zeros appear in dark and higher values are lightened. As we can observe in Figure \ref{generations}, this qualitative evaluation clearly shows that the BVAE generates a rubber compound dataset presenting much more realistic patterns than the one produced by the GAN. This qualitative observation is confirmed by quantitative metrics (cf Figure \ref{fid_inception}). By using this kind of visual verification, one can easily discriminate between models and quickly spot poor generations. However, selecting the best generative model needs finer evaluation metrics as the ones developed in the following.
\begin{figure}[t]
	\begin{center}
		\includegraphics[width=0.65\textwidth]{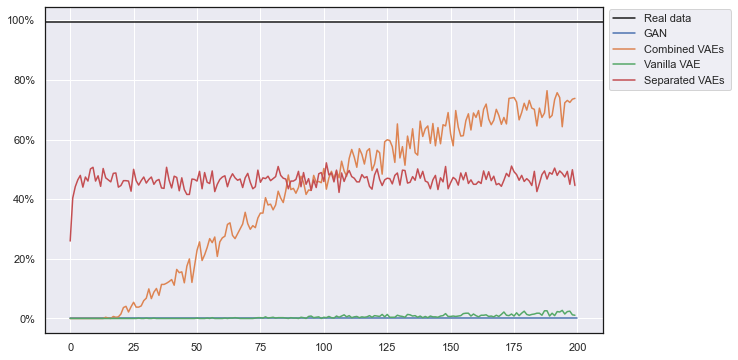}
		\caption{Evolution of the ERA scores by epoch. The BVAE is able to get closer to the real data score, whereas the GAN and Vanilla VAE perform very poorly with a 0\% score for the GAN.}
		\label{ratio}
	\end{center}
\end{figure}
\textbf{FID* and KL scores.} 
Figure \ref{fid_inception} presents the evolution of FID* and KL score through epochs for the different generative models.
The two scores are computed on 10 000 generations. The KL score is computed by averaging the KLs obtained for the three properties. In addition we plot the test scores computed between two subsets of the real data, and the noise scores computed between the real data and a random Gaussian noise.

We can observe that the GAN performs poorly compared to the other models and is unstable during training, but sill has generally lower scores than the random noise. We can also observe that the two metrics are coherent and give the same score ranking for all models. We notice that the Unbinded-VAE model has better scores in the first epochs before being overrun by the BVAE, which is due to the fact that for the Unbinded-VAE, the mask generation is already pretrained.

The most notable point is the improved performance of the BVAE against the Unbinded-VAE. This fact shows that the learning of the ratios by the Ratio-VAE has a positive influence on the learning of the mask by the Mask-VAE. 
Indeed, some parts of the mask might be more important than others. For instance if a component’s ratio is always high when this component is present in the compound, a good prediction of the mask for this component will be more valuable than a good mask prediction for a component that always appears with small ratios.


\textbf{"Expert Rules Agreement" score.} Looking at the ERA score is even more instructive. We observe that only the Unbinded-VAE and the BVAE models provide generated data that match all the rules that a rubber compound must at least verify to be considered plausible. It means that the data produced by the GAN and the Vanilla VAE would not be considered as valid rubber compound by experts. This highlights one limitation of the FID* and the KL divergence. Indeed, according to these metrics Vanilla VAE was close from the two others but as shown here, only 1\% of the data generated can be considered as realistic. 

\section{Conclusion}

The poor performances of standard generative models like GANs or VAEs applied to industrial material generations such as rubber compound, makes them unusable and not very popular amongst industrial experts. This work proposes a VAE-based generative model able to overcome the difficulties
encountered by standard generative models, and to generate plausible industrial compound-like data. We believe that this work should contribute to the popularization of deep generative methods in the industry, and demonstrates the potential that these methods can bring to this field.

\section*{Acknowledgments}
This research was funded by Manufacture Française des Pneumatiques \textbf{Michelin} and the Industrial Data Analytics and Machine Learning chair of \textbf{Centre Borelli} from \textbf{ENS Paris Saclay}.

\bibliographystyle{plain}
\typeout{}
\bibliography{neurips_2021}

\begin{thebibliography}{10}

\bibitem{chakraborti2004genetic}
N~Chakraborti.
\newblock Genetic algorithms in materials design and processing.
\newblock {\em International Materials Reviews}, 49(3-4):246--260, 2004.

\bibitem{dan2020generative}
Yabo Dan, Yong Zhao, Xiang Li, Shaobo Li, Ming Hu, and Jianjun Hu.
\newblock Generative adversarial networks (gan) based efficient sampling of
  chemical composition space for inverse design of inorganic materials.
\newblock {\em npj Computational Materials}, 6(1):1--7, 2020.

\bibitem{gomez2018automatic}
Rafael G{\'o}mez-Bombarelli, Jennifer~N Wei, David Duvenaud, Jos{\'e}~Miguel
  Hern{\'a}ndez-Lobato, Benjam{\'\i}n S{\'a}nchez-Lengeling, Dennis Sheberla,
  Jorge Aguilera-Iparraguirre, Timothy~D Hirzel, Ryan~P Adams, and Al{\'a}n
  Aspuru-Guzik.
\newblock Automatic chemical design using a data-driven continuous
  representation of molecules.
\newblock {\em ACS central science}, 4(2):268--276, 2018.

\bibitem{goodfellow2014generative}
Ian Goodfellow, Jean Pouget-Abadie, Mehdi Mirza, Bing Xu, David Warde-Farley,
  Sherjil Ozair, Aaron Courville, and Yoshua Bengio.
\newblock Generative adversarial nets.
\newblock {\em Advances in neural information processing systems}, 27, 2014.

\bibitem{gui2020review}
Jie Gui, Zhenan Sun, Yonggang Wen, Dacheng Tao, and Jieping Ye.
\newblock A review on generative adversarial networks: Algorithms, theory, and
  applications.
\newblock {\em arXiv preprint arXiv:2001.06937}, 2020.

\bibitem{guimaraes2017objective}
Gabriel~Lima Guimaraes, Benjamin Sanchez-Lengeling, Carlos Outeiral, Pedro
  Luis~Cunha Farias, and Al{\'a}n Aspuru-Guzik.
\newblock Objective-reinforced generative adversarial networks (organ) for
  sequence generation models.
\newblock {\em arXiv preprint arXiv:1705.10843}, 2017.

\bibitem{heusel2017gans}
Martin Heusel, Hubert Ramsauer, Thomas Unterthiner, Bernhard Nessler, and Sepp
  Hochreiter.
\newblock Gans trained by a two time-scale update rule converge to a local nash
  equilibrium.
\newblock {\em Advances in neural information processing systems}, 30, 2017.

\bibitem{jadon2020survey}
Shruti Jadon.
\newblock A survey of loss functions for semantic segmentation.
\newblock In {\em 2020 IEEE Conference on Computational Intelligence in
  Bioinformatics and Computational Biology (CIBCB)}, pages 1--7. IEEE, 2020.

\bibitem{karras2019style}
Tero Karras, Samuli Laine, and Timo Aila.
\newblock A style-based generator architecture for generative adversarial
  networks.
\newblock In {\em Proceedings of the IEEE/CVF Conference on Computer Vision and
  Pattern Recognition}, pages 4401--4410, 2019.

\bibitem{kingma2014adam}
Diederik~P Kingma and Jimmy Ba.
\newblock Adam: A method for stochastic optimization.
\newblock {\em arXiv preprint arXiv:1412.6980}, 2014.

\bibitem{Kingma2014AutoEncodingVB}
Diederik~P. Kingma and M.~Welling.
\newblock Auto-encoding variational bayes.
\newblock {\em CoRR}, abs/1312.6114, 2014.

\bibitem{krizhevsky2012imagenet}
Alex Krizhevsky, Ilya Sutskever, and Geoffrey~E Hinton.
\newblock Imagenet classification with deep convolutional neural networks.
\newblock {\em Advances in neural information processing systems},
  25:1097--1105, 2012.

\bibitem{kusner2017grammar}
Matt~J Kusner, Brooks Paige, and Jos{\'e}~Miguel Hern{\'a}ndez-Lobato.
\newblock Grammar variational autoencoder.
\newblock In {\em International Conference on Machine Learning}, pages
  1945--1954. PMLR, 2017.

\bibitem{lameijer2006molecule}
Eric-Wubbo Lameijer, Joost~N Kok, Thomas B{\"a}ck, and Ad~P IJzerman.
\newblock The molecule evoluator. an interactive evolutionary algorithm for the
  design of drug-like molecules.
\newblock {\em Journal of chemical information and modeling}, 46(2):545--552,
  2006.

\bibitem{maziarka2020mol}
{\L}ukasz Maziarka, Agnieszka Pocha, Jan Kaczmarczyk, Krzysztof Rataj, Tomasz
  Danel, and Micha{\l} Warcho{\l}.
\newblock Mol-cyclegan: a generative model for molecular optimization.
\newblock {\em Journal of Cheminformatics}, 12(1):1--18, 2020.

\bibitem{shamir2019continuous}
Reuben~R Shamir, Yuval Duchin, Jinyoung Kim, Guillermo Sapiro, and Noam Harel.
\newblock Continuous dice coefficient: a method for evaluating probabilistic
  segmentations.
\newblock {\em arXiv preprint arXiv:1906.11031}, 2019.

\bibitem{Sohn2015CVAE}
Kihyuk Sohn, Honglak Lee, and Xinchen Yan.
\newblock Learning structured output representation using deep conditional
  generative models.
\newblock {\em Advances in neural information processing systems},
  28:3483--3491, 2015.

\bibitem{Szegedy2016InceptionNetwork}
C~Szegedy, V~Vanhoucke, S~Ioffe, J~Shlens, and ZB~Wojna.
\newblock Rethinking the inception architecture for computer vision.
\newblock In {\em 2016 IEEE Conference on Computer Vision and Pattern
  Recognition}, volume 2016, pages 2818--2826. IEEE, 2016.

\bibitem{vahdat2020nvae}
Arash Vahdat and Jan Kautz.
\newblock Nvae: A deep hierarchical variational autoencoder.
\newblock {\em arXiv preprint arXiv:2007.03898}, 2020.

\bibitem{zhu2017unpaired}
Jun-Yan Zhu, Taesung Park, Phillip Isola, and Alexei~A Efros.
\newblock Unpaired image-to-image translation using cycle-consistent
  adversarial networks.
\newblock In {\em Proceedings of the IEEE international conference on computer
  vision}, pages 2223--2232, 2017.

\end{thebibliography}

\newpage
\appendix

\section{Appendices}
\subsection{Architectures and hyper-parameters}


\label{architectures}

\begin{table}[H]
    \centering
	\small
    \begin{tabular}{|l|c|c|c|c|c|}
    \hline\hline
       	Hyper-parameter  &	Mask-VAE&	Ratio-VAE&	Vanilla VAE&	GAN\\
        \hline
	    Latent dimension & 64	& 32 & 128	& 1024\\
        \hline
	    Learning rate  & 0.001	& 0.01	& 0.001	& 0.001\\
        \hline
        $\lambda $&  1 & 10	& -	& -		 \\
        \hline
	    $\gamma$ & \multicolumn{2}{l|}{5 \small{only concerns the BVAE}}  & -	& -		\\
	    \hline
	    $\beta$  & \multicolumn{2}{l|}{10 \small{only concerns the BVAE}}  & -	& -	\\
        \hline
	
    \end{tabular}
    \caption{Selected hyper-parameters. Exceptionally for the GAN's latent dimension we went up to 1024, exceeding the gridsearch rang limit  due to its poor generations quality.}
    \label{tab: hyperparameters}
\end{table}

\begin{table}[H]
    \centering
	\small
    \begin{tabular}{|l|l|l|}
    \hline\hline
       Model  &	Layer size & Activation\\
        \hline
	\multirow{4}{*}{Encoder} & 512 & Relu \\ \cline{2-3}
	& 1024 & Relu\\ \cline{2-3}
	& 256 & Relu\\ \cline{2-3}
	& $2\times$latent dim & Linear\\ 
 	\hline	
       \multirow{4}{*}{Decoder} & 512 & Relu \\ \cline{2-3}
	& 1024 & Relu\\ \cline{2-3}
	& 256 & Relu\\ \cline{2-3}
	& 99  & Sigmoid\\ 
	\hline
	Loss function & \multicolumn{2}{l|} {$\mathcal{L}_{\text{VAE}}$ for the Vanilla VAE and $\mathcal{L}_{\text{VAE}_M}$ for Mask-VAE } \\
	
    \hline
    \end{tabular}
    \caption{Vanilla VAE and Mask-VAE architecture}
    \label{tab:vae1}
\end{table}

\begin{table}[H]
    \centering
	\small
    \begin{tabular}{|l|l|l|}
    \hline\hline
       Model  &	Layer size & Activation\\
        \hline
	\multirow{3}{*}{Encoder} & 256 & Relu \\ \cline{2-3}
	& 256 & Relu\\ \cline{2-3}
	& $2\times$latent dim & Linear\\ 
 	\hline	
       \multirow{2}{*}{Decoder} & 256 & Relu \\ \cline{2-3}
	& 256 & Relu\\ \cline{2-3}
	& 99 & Sigmoid\\
	\hline
	Loss function & \multicolumn{2}{l|} {$\mathcal{L}_{\text{VAE}_R}$ } \\
	
    \hline
    \end{tabular}
    \caption{Ratio-VAE architecture}
    \label{tab:vae2}
\end{table}
\begin{table}[H]
    \centering
	\small
    \begin{tabular}{|l|l|l|}
    \hline\hline
       Model  &	Layer size & Activation\\
        \hline
	\multirow{4}{*}{Generator} & 512 & LeakyReLU($\alpha$=0.2) \\ \cline{2-3}
	& 1024 & LeakyReLU($\alpha$=0.2)\\ \cline{2-3}
	& 256 & LeakyReLU($\alpha$=0.2)\\ \cline{2-3}
	& 99& Linear\\ 
 	\hline	
       \multirow{4}{*}{Discriminator} & 512 & LeakyReLU($\alpha$=0.2) \\ \cline{2-3}
	& 1024 & LeakyReLU($\alpha$=0.2)\\ \cline{2-3}
	& 256 & LeakyReLU($\alpha$=0.2)\\ \cline{2-3}
	& 1 & Sigmoid\\ 
	\hline
	Loss function & \multicolumn{2}{l|} {$\text{BCE}$ } \\
	
    \hline
    \end{tabular}
    \caption{GAN architecture}
    \label{tab:gan}
\end{table}
\end{document}